\documentclass{article}

\usepackage{PRIMEarxiv}

\usepackage[utf8]{inputenc} % allow utf-8 input
\usepackage[T1]{fontenc}    % use 8-bit T1 fonts
\usepackage{hyperref}       % hyperlinks
\usepackage{url}            % simple URL typesetting
\usepackage{booktabs}       % professional-quality tables
\usepackage{amsfonts}       % blackboard math symbols
\usepackage{nicefrac}       % compact symbols for 1/2, etc.
\usepackage{microtype}      % microtypography
\usepackage{lipsum}
\usepackage{fancyhdr}       % header
\usepackage{graphicx}       % graphics
\graphicspath{{media/}}     % organize your images and other figures under media/ folder

\usepackage{float}

\restylefloat{figure}
\restylefloat{table}
\usepackage{algorithm}
\usepackage{algorithmicx}
\usepackage{algpseudocode}
\usepackage{multirow}

\usepackage{mathtools,amssymb}
\title{VGFL-SA: Vertical Graph Federated Learning Structure Attack Based on Contrastive Learning
}

\author{
  Yang Chen, Bin Zhou \\
chenyang2753@sdtbu.edu.cn} 
\begin{document}
\maketitle

\begin{abstract}
Graph Neural Networks (GNNs) have gained attention for their ability to learn representations from graph data.   Due to privacy concerns and conflicts of interest that prevent clients from directly sharing graph data with one another, Vertical Graph Federated Learning (VGFL) frameworks have been developed. Recent studies have shown that VGFL is vulnerable to adversarial attacks that degrade performance. However, it is a common problem that client nodes are often unlabeled in the realm of VGFL. Consequently, the existing attacks, which rely on the availability of labeling information to obtain gradients, are inherently constrained in their applicability. This limitation precludes their deployment in practical, real-world environments. To address the above problems, we propose a novel graph adversarial attack against VGFL, referred to as VGFL-SA, to degrade the performance of VGFL by modifying the local clients structure without using labels. Specifically, VGFL-SA uses a contrastive learning method to complete the attack before the local clients are trained. VGFL-SA first accesses the graph structure and node feature information of the poisoned clients, and generates the contrastive views by node-degree-based edge augmentation and feature shuffling augmentation.  Then, VGFL-SA uses the shared graph encoder to get the embedding of each view, and the gradients of the adjacency matrices are obtained by the contrastive function. Finally, perturbed edges are generated using gradient modification rules. We validated the performance of VGFL-SA by performing a node classification task on real-world datasets, and the results show that VGFL-SA achieves good attack effectiveness and transferability.
\end{abstract}

% keywords can be removed
\keywords{	Graph Adversarial Attacks; Contrastive Learning; Vertical Federated Learning; Unsupervised Learning
}

\section{Introduction}\label{Section:Intro}

In recent years, deep learning models based on graph data (Graph Neural Networks, GNNs) have been widely used in various real-world domains, for example, medical information \cite{WOS:001235030400001, WOS:001185440500001}, bioinformation \cite{WOS:001263692800047,WOS:001217577600001} and social networks \cite{WOS:001161624500003,WOS:001175221000031,WOS:001252433900001}. GNNs apply deep learning-based methods on graph data to learn the structural and feature information by aggregating the neighboring information of the nodes, and achieve superior performance in node classification \cite{WOS:001261438100001,WOS:001260907100001}, link prediction \cite{WOS:001310810700001,WOS:001263692800047}, graph embedding \cite{WOS:001324833600001,WOS:001167452200006}.

GNNs have achieved substantial advancements in the domain of graph data learning \cite{ WOS:001236624200001,WOS:001242386500001}. 
Nonetheless, a predominant characteristic among these methodologies is the collective storage of graph data.
In real life, real graph data is usually decentralized, i.e., the data is usually owned by different clients (also known as data owners) \cite{WOS:001360494400010}. Due to privacy concerns, clients tend to protect the original data (especially sensitive and private data), leading to the data silo problem \cite{ WOS:001224177900008}. For example, GNNs are expected to comprehensively assess a patient's health status and help the patient to predict potential diseases or find out the causes of the diseases. In the healthcare domain, different hospitals or healthcare organizations usually have a large amount of patient data, which is not only private but also often stored in the form of graph structures. Hospitals can use medical images to diagnose diseases, where each patient's medical records and images can be viewed as nodes in the graph, and collaboration between hospitals can be considered as edges of the graph. Hospitals usually cannot share patient-specific information due to data privacy protection and legal and regulatory restrictions. Therefore, cross-hospital collaboration for federated training to develop more powerful disease prediction models is still a great need.

In order to solve the above problems, the researchers proposed the Graph Federated Learning (GFL) framework \cite{WOS:001253867000023}. GFL effectively solves the data isolation problem through collaborative training, especially in scenarios involving data privacy protection among multiple data owners (e.g., hospitals, banks) \cite{WOS:001294329300001}. Specifically, each data owner (e.g., individual hospitals, different healthcare organizations) trains GNNs independently on its own local data. Since graph data itself has the features of nodes and edges (e.g., social networks, knowledge graphs, patient relationships in medical images), each client uses its local graph data to train the model and extracts the relational information and structural features among nodes \cite{WOS:001252500000001}. During the training process, the clients do not share the raw data directly, but only share the parameters (e.g., weights and gradients) of the local models or other models update information. Clients upload their local model parameters to the central server or aggregation node for global model update \cite{WOS:001175221000057}.
In particular, after receiving model parameters from multiple clients, the central server uses algorithms such as federated averaging to aggregate the updates of each local model to generate a globally shared model. This process does not require the exchange of any local data and guarantees privacy protection \cite{DBLP:journals/sigkdd/FuZDCL22}. The aggregated global model is fed back to individual clients, who can continue to train and fine-tune it locally. This process can be repeated for multiple rounds until the model converges.

\begin{figure*}[h]
	\centering
	\includegraphics[width=1\textwidth]{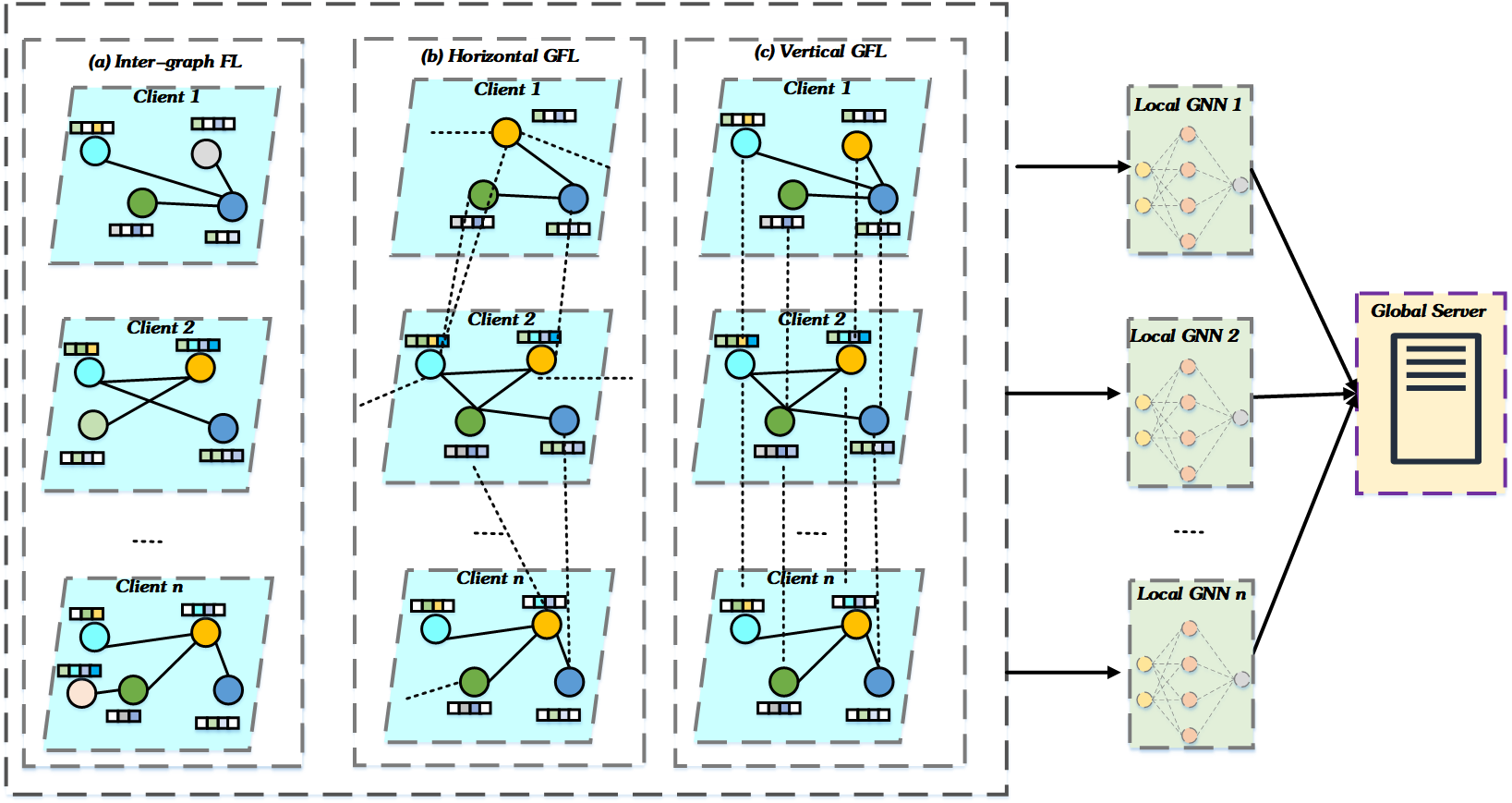}
	\caption{Diagram of the three GFL frameworks. A GFL model consisting of $n$ clients, the global server gets the final global node embeddings by collecting the node embeddings from each client, and then completes the downstream tasks. (a) In Inter-graph FL, each client has independent graph data, and the data between each client is independent. (b) In Horizontal GFL, each client owns a graph that is similar in the type and features of nodes, but the number of nodes is different, and the dotted lines indicate the missing edges. (c) In Vertical GFL, each client owns the same nodes, but the nodes have different features.}\label{fig1}
\end{figure*}

According to the way graph data is distributed among clients, GFL is categorized as follows (1). Inter-graph Federated Learning (Inter-graph FL), (2). Intra-graph Federated Learning (Intra-graph FL) \cite{GFL}. Fig. \ref{fig1}  (a) shows the Inter-graph FL framework, which is concerned with collaborative learning between different graphs \cite{2022FedGCN}. Each client has independent graph data (e.g., different communities in a social network, data from different hospitals, data from multiple organizations), and these graphs can have different nodes, edges, and structures. The structure of the graph data may different greatly from one participant to another, leading to difficulties when collaborating across graphs. Most of the existing GFL focuses with Intra-graph FL. Among them, Intra-graph FL can be categorized into Horizontal GFL and Vertical GFL. Horizontal Graph Federated  Learning (HGFL) is usually applied when different clients hold similar types of graphs, i.e., the graphs held by each participant are similar in terms of node types and features, but with different node numbers \cite{DBLP:journals/tist/YangLCT19}, as shown in Fig. \ref{fig1} (b). 
In the Vertical Graph Federated  Learning (VGFL), different clients have information about different features of the same set of entities \cite{DBLP:journals/corr/abs-1909-12946,DBLP:conf/jist/ChenZYJC21}, as shown in Fig. \ref{fig1} (c). This is very common in real-world scenarios, e.g., in financial systems where users usually have financial records in different financial institutions and data is vertically categorized in different financial institutions. Another real-world example is, in an e-commerce VGFL, one client have information about the user's social relationships, and another client has information about the user's consumption behavior.

GFL helps to keep local raw data private and to train collaboratively with distributed clients, it still suffers from the problem of vulnerability.
However, most of the current research on GFL focuses on the performance and ignores the robustness under adversarial attacks \cite{GraphFraudster}.
In fact, GFL is susceptible to adversarial attacks such as modifying graph structures, node features or embeddings.
Adding or removing a small number of edges in the client's graph will produce poisoned node embeddings, leading to performance degradation when the embeddings are uploaded to the central server. With the deployment of GFL in real production environments, this vulnerability has raised many concerns in both academia and industry, e.g.,  an attacker may bypass the detection system and obtain a higher credit value by establishing a connection to a high-credit client in financial systems.

In recent years, a number of works have been proposed against adversarial attacks on graph \cite{WOS:001196644600025, WOS:001196644600025,WOS:001261404700004}. Researchers have explored less about GFL robustness, only Graph-Fraudster \cite{GraphFraudster}. Specifically, Graph-Fraudster uses graph structure, node features and labels information to get the gradient information of nodes, the aim of adding noise to the global node embedding to generate adversarial perturbations. The premise of Graph-Fraudster implementation is that the attack is semi-supervised, i.e., node labels information is necessary to perform the attack. However, the label information of many nodes is often scarce in real life, especially in VGFL. Due to the decentralized nature of the data many nodes are unlabeled. A core goal of VGFL is to protect data privacy, where node labels information cannot be shared among different clients. In addition, labeling nodes usually takes a lot of time and resources, especially in large-scale networks. As a result, Graph-Fraudster will be ineffective in environments where node label information.

To address the above problems, we propose a \textbf{\textit{V}}ertical \textbf{\textit{G}}raph \textbf{\textit{F}}ederated \textbf{\textit{L}}earning \textbf{\textit{S}}tructure \textbf{\textit{A}}ttack Based on contrastive learning (VGFL-SA), which aims to modify the graph structure to complete the attack without using the information of node labels, degrading the performance of the downstream tasks of the VGFL.
VGFL unintentionally conditions the success of adversarial attacks due to data bias between clients. In other words, the graph structure is different for different clients, it is not easily detectable by the defense model when the attack modifies the structure. Specifically, VGFL-SA modifies the graph structure to generate a perturbed graph to complete the attack before the local client is trained. VGFL-SA chooses to attack a particular client, and after obtaining the graph structure and the node features, it uses data enhancement to obtain comparative views. We use the node degree characteristic to generate edge augmentation, nodes with larger node degree values contain more information, and rewiring at such nodes tends to be more efficient. In order to avoid random feature augmentation that would generate features that have not been seen in the original feature space. We use feature shuffling for feature augmentation, which does not rewrite the node features. Then the embedding of each view is obtained using a shared graph encoder, and the gradient of the adjacency matrix is obtained through a contrastive function. Finally, VGFL-SA flips the edges with maximum gradient using the modification rule to get the client's poisoned graph. Numerous experiments have shown that VGFL-SA can outperform existing unsupervised baselines and even outperform supervised attacks in many metrics.

Finally, we summarize the contributions of this paper as follows:

$\bullet$ We propose an unsupervised attack against VGFL, referred to as VGFL-SA. VGFL-SA does not use node labels to generate structural perturbations that degrade the VGFL performance of downstream tasks.

$\bullet$ We use graph comparative learning method to obtain the gradient of the adjacency matrix. We use node degree based edge augmentation and feature shuffling augmentation to reduce the randomness of augmentation improved the performance of comparative learning.

$\bullet$ We demonstrate the performance of VGFL-SA on three datasets comparing semi-supervised learning based attacks and unsupervised learning based attacks respectively. The results indicate that VGFL-SA outperforms the newest unsupervised learning attacks and has comparable performance with some semi-supervised attack methods in some metrics.

The rest of this work is organized as follows. 
Section \ref{S2}  reviews the work related to this paper, including graph federation learning, graph comparative learning and graph adversarial attacks. In section \ref{S3}, we explain the fundamental knowledge related to this paper, including GNNs and GVFL definitions and the attack environment of VGFL-SA. In section \ref{S4}, we introduce VGFL-SA in detail, including graph augmentation, gradient attack module, and give the algorithmic framework and complexity. 
Section \ref{S5} shows the attack setup and experimental results of VGFL-SA. Finally,  Section \ref{Sec:conclusion} gives the conclusion of this paper.

\section{Related Work}\label{S2}
In this section, we briefly summarize existing works on graph federation learning, graph contrastive learning, and graph adversarial attack.

\subsection{ Graph Federation Learning}\label{HypergraphLearning}

GFL trains models across multiple data owners without exchanging the original data and can effectively protect user privacy \cite{DBLP:journals/pvldb/LiWZZLW23}. Ni et al. \cite{DBLP:journals/corr/abs-2106-11593} proposed a federated privacy-preserving node GCN learning framework (FedVGCN), which is suitable for the case where data is vertically distributed. FedVGCN splits the computational graph data into two parts. For each iteration of training, both sides pass intermediate results under homomorphic encryption. The literature \cite{DBLP:conf/ijcai/0001ZZWLWWLWZ22} proposed vertical federated graph neural network (VFGNN), VFGNN keeps the private data (i.e., edges, node features, and labels) on the clients, and the rest of the information is given to be uploaded to a semi-honest server for training. Liu et al. \cite{DBLP:conf/apweb/LiuZXSSHDK23} proposed a federated learning of subgraphs with global graph reconstruction (FedGGR).  For the data silo problem, Zhang et al. \cite{DBLP:journals/tkdd/ZhangMCCSZ24} proposed FedEgo, FedEgo uses GraphSAGE on ego-graphs to fully exploit the structural information and Mixup to address the privacy issues. Xue et al. \cite{DBLP:journals/tbd/XueZJT24} proposed a new framework for federated learning of personality graphs based on variational graph self-encoders (FedVGAE). Du et al. \cite{DBLP:journals/kbs/DuLLDH24} proposed a new efficient GFL framework (FedHGCN), FedHGCN is able to be co-trained in high-dimensional space to obtain graph-rich hierarchical features. In addition, FedHGCN uses a node selection strategy to remove nodes with redundant information from the graph representation to improve efficiency.  Huang et al. \cite{DBLP:journals/tbd/HuangLLJWH23} proposed a federated learning cross-domain knowledge graph embedding model (FedCKE) in which entity/relationship embeddings between different domains can interact securely without data sharing. Zheng et al. \cite{DBLP:journals/tii/LiBXJC23} proposed a cross-firm recommendation GNNs training framework (FL-GMT), which no longer uses traditional federation learning training methods (e.g., averaged federation), and designs a loss-based federation aggregation algorithm to improve the sample quality. The literature \cite{DBLP:journals/tist/LiuHZZXC22} proposed a federated multi-task graph learning (FMTGL) framework to address issues in privacy preserving and scalable schemes.

\subsection{Graph Contrastive Learning}\label{2.2}
Graph contrastive learning is an unsupervised learning method that aims to learn effective representations from graph data \cite{DBLP:journals/ijon/YuJ24}.
Meng et al. \cite{DBLP:conf/cikm/MengL23} proposed an informative contrastive learning (IMCL) which uses a graph augmentation generator for distinguishing the augmented view from the original view. Besides, IMCL uses a pseudo-label generator to generate pseudo-labels as a supervisory signal to ensure that the results of the augmented view classification are consistent with the original view.  Feng et al. \cite{DBLP:journals/tkdd/FengJZT24} proposed the ArieL method, which introduces an adversarial graph view for data augmentation and also uses information regularization methods for stable training. In addition, ArieL uses subgraph sampling to extend to different graphs. Jiang et al. \cite{DBLP:journals/nn/JiangB24} proposed probabilistic graph complementary contrastive learning (PGCCL) for adaptive construction of complementary graphs, which employs a Beta mixture model to distinguish intraclass similarity and interclass similarity, and solves the problem of inconsistent similarity distributions of data. Yang et al. \cite{DBLP:journals/tkde/YangWZCYLX24} proposed a graph knowledge contrastive learning (GKCL), which uses exploits multilevel graph knowledge to create noise-free contrastive views that can alleviate the problem of introducing noise and generating samples that require additional storage space during graph augmentation. The literature \cite{DBLP:journals/nn/LiangDZM0G23} proposed an implicit graph contrastive learning (IGCL), which avoids the situation where changing certain edges or nodes may accidentally change the graph features by reconfiguring the topology of the graph. Li et al. \cite{DBLP:journals/jksucis/LiMYXCZ24} proposed a line graph contrastive learning (Linegcl), the core of which is to transform the original graph into the corresponding line graph, solving the deficiencies of the existing methods in understanding the overall features and topology of the graph. Since the similarity-based methods are defective in terms of node information loss and similarity metric generalization ability.
The literature \cite{DBLP:journals/pr/ZhangSMZ23} proposed a linear graph contrastive learning (LGCL), which obtains subgraph views by sampling h-hop subgraphs of target node pairs, and then maximizes mutual information after transforming the sampled subgraphs into linear graphs. The literature \cite{DBLP:journals/tnn/PengCTLW24} proposed a dyadic contrastive learning network (DCLN), which is based on a self-supervised learning approach to enhance the model performance through the pairwise reduction of redundant information about the learned latent variables.

\subsection{Graph Adversarial Attack}\label{2.3}

GNNs have achieved significant success in many domains and are vulnerable to adversarial attacks due to their high dependence on graph structure and node features. Graph Adversarial Attacks are defined as small modifications to the input graph that can cause GNNs to output incorrect predictions or classification results \cite{WOS:001293902200001}.
The literature \cite{DBLP:journals/tkde/WangCXBZWZZ24} proposed a generalized attack framework (CAMA) that generates the importance of nodes through graph activation mapping and its variants.
Zhang et al. \cite{DBLP:conf/europar/ZhangLWYYYF24} proposed the first framework for training adversarial attacks on distributed GNNs (Disttack), which centers on disrupting the gradient synchronization between computational nodes by injecting adversarial attacks into individual computational nodes. Ennadir et al. \cite{DBLP:conf/complexnetworks/EnnadirANVB23} proposed a model for generating adversarial perturbations by generating entirely new nodes (UnboundAttack), which uses advances in graph generation to generate subgraphs. Zhao et al. \cite{DBLP:journals/tkde/ZhaoYYXZLLC24} proposed a new gradient-based attack method for the robustness of dynamic graph neural networks from an optimization point of view, which centers on using gradient dynamics to attack the structure of the graph. Zhu et al. \cite{DBLP:journals/corr/abs-2308-07834} proposed a partial graph attack (PGA) which uses a hierarchical target selection strategy that allows the attacker to focus only on vulnerable nodes. Then, the optimal perturbed edges are selected by a cost-effective node selection strategy. Aburidi et al. \cite{DBLP:conf/eais/AburidiM24} proposed an attack based on training time optimization, which first optimizes the graph as hyperparameters and then uses convex relaxation and projected momentum optimization techniques to generate structural attacks. Since existing attackers need to access the target model without considering the budget allocation, the literature \cite{DBLP:journals/kbs/CaoCY24} proposed a targeted labeling attack (ETLA), which allocates the attack budget in terms of both the search space and the optimized target, allowing the attack to achieve the best performance. The literature \cite{DBLP:journals/corr/abs-2401-02663} proposed a backdoor attack for the link prediction task that uses individual nodes as triggers and selects poisoned pairs of nodes in a training graph, and then embeds the backdoor in the training process of their GNNs.

\section{Preliminaries}\label{S3}

This section first introduces the knowledge related to graphs. Then, definitions related to GNNs and VGFL are elaborated and formalized. Finally, the attack environment of VGFL-SA is introduced.

For convenience, Table \ref{notations} gives the frequently used notations.

\begin{table}[!ht]
	\begin{center}
		\caption{Notations frequently used in this paper and their corresponding descriptions.}\label{notations}%
		\begin{tabular}{cc}
			\toprule
			Notation & Description\\
			\midrule
			$G$& Clean graph dataset\\
			${G^{\prime}}$& Poisoned graph dataset\\
			$V$& Set of nodes of the clean graph\\
			$n$& Number of nodes\\
			$X$& Feature matrix of the graph graph\\
			$E$& Set of edges of the clean graph\\
			$m$ &Number of edges\\
			$E^{\prime}$&  Set of edges of of the perturbed graph\\
			$A$& Adjacency matrix of the clean graph\\
			$A^{\prime}$& Adjacency matrix of the perturbed graph\\
			$Y$&True label\\
			$\widehat Y$&Prediction label\\
			$\Delta$ & Attack budget\\	
			$\alpha $ & Budget factor\\
			$K $ & Number of clients\\
			$K^{\prime} $ & Number of poisoned clients\\
			$S$ & Central server\\
			\midrule
		\end{tabular}
	\end{center}
\end{table}

\subsection{Graph Definition}\label{3.1}

Given a attribute graph ${\rm{G = }}(V,E,X)$, where $V = \{ {v_1},{v_2},...,{v_n}\} $ represents the set of nodes, $n$ denotes the number of nodes, $E = \{ {e_1},{e_2},...,{e_m}\} $ is the set of edges, $m$ is the number of edges, and $X \in {\mathbb{R}^{n \times d}}$ represents the set of node features, where $d$ denotes the dimension of the features. We use the adjacency matrix $A \in {\{ 0,1\} ^{n \times n}}$ to represent the link relationship between nodes. When there is a link between node $i$ and node $j$, there ${A_{ij}} = 1$.
Conversely, ${A_{ij}} = 0$ indicates that no link exists between node $i$ and node $j$.

\subsection{Graph Neural Networks}\label{3.2}

In recent years, more and more graph deep models have been proposed \cite{DBLP:conf/nips/HamiltonYL17,WOS:001369501700001}. Among them, the state-of-the-art Graph Convolutional Network (GCN) \cite{GCN}, Graph Attention Network (GAT) \cite{GAT} have achieved excellent performance in node classification, link prediction and graph classification tasks.

\textbf{GCN}: GCN is a graph neural network based on the idea of convolutional neural network, which aims to learn the representation of graph nodes. The core  is to perform convolutional operations on node features through the adjacency matrix of the graph, so as to realize the propagation and aggregation of information.
Specifically, GCN performs the weighted average of the features of each node and its neighboring nodes. The update of the node features is computed through the adjacency matrix of the graph, and the update is performed by graph convolution operation as follows:

\begin{equation}
	{H^{(l + 1)}} = \sigma \left( {\hat A{H^{(l)}}{W^{(l)}}} \right).
\end{equation}

${H^{(l)}}$ is the feature output matrix of the $l$-th layer. $\hat A = A + {I_n}$ is the normalized adjacency matrix of the graph (usually with the addition of the self-loop) and ${I_n}$ is the unit matrix. ${W^{(l)}}$ is the learned weight matrix. $\sigma $ is the activation function (e.g. ReLU). GCN aggregates the node's local neighborhood information to the node itself, enabling each node's representation to contain information about its neighbors.

\textbf{GAT}: GAT introduces an attention mechanism to adaptively aggregate information from neighboring nodes by dynamically computing the importance (weight) of different neighboring nodes. Unlike GCN, which uses a fixed adjacency matrix to weight neighboring nodes, GAT determines the influence of each neighbor through the attention weights among nodes.

GAT uses the attention mechanism to compute the relationship weights between nodes. For node $i$ and node $j$, GAT calculates the attention weights between them and then aggregates the information of neighboring nodes based on these weights. The specific attention weight calculation formula is as follows:

\begin{equation}
	{\alpha _{ij}} = \frac{{\exp \left( {{\rm{LeakyReLU}}\left( {{{{a}}^T}\left[ {{{W}}{{{h}}_i}|{{W}}{{{h}}_j}} \right]} \right)} \right)}}{{\sum\limits_{k \in {{\cal N}_{(i)}}} {\exp \left( {{\rm{LeakyReLU}}\left( {{{{a}}^T}\left[ {{{W}}{{{h}}_i}|{{W}}{{{h}}_k}} \right]} \right)} \right)} }}.
\end{equation}

${{{h}}_i}$ and ${{{h}}_j}$ are the feature vectors of node $i$ and node $j$, respectively. ${W}$ is a learned weight matrix.  ${a}$ is a learned attention parameter that will be dynamically adjusted at each node of the graph for its neighboring nodes. The feature vector of node $i$ can be expressed as:

\begin{equation}
	{{h_i}} = \sigma (\sum\limits_{j \in {{\cal N}_{(i)}}} {{\alpha _{ij}}W}{h_j} ).
\end{equation}

\subsection{ Vertical Graph Federated Learning}\label{3.2}
VGFL is an approach that combines graph data and vertical federation learning. In this framework, multiple clients learn collaboratively based on parts of the graph data held by each of them without sharing the original data. Each client holds different node features or a part of the graph, and the goal is to learn a global graph model by cross-party collaborative training. Suppose that in VGFL, there is a collection of clients and a server $S$, where $K$ is the number of clients. Each client holds in each client a part of the data of the graph $G$, i.e., ${G_i} = (V,{E_i},{X_i})$. Where nodes are shared in each client, but node features and node adjacencies are different, i.e., in any client $i$ and client $j$, ${E_i} \cap {E_j} = \emptyset $, ${X_i} \cap {X_j} = \emptyset $. The sum of the data of all clients is $\sum\limits_{i = 1}^K {{E_i}}  = E$ , $\sum\limits_{i = 1}^K {{X_i}}  = X$.

In VGFL, each client trains local GNNs with its private data and updates the embedding for server aggregation.

\begin{equation}
	{h_{global}} \leftarrow CONCAT({h_1},...,{h_i},...,{h_K}), \quad  {\rm{ s}}{\rm{.t.}} \quad {{h}_i} = f_\theta ^i({A_i},{X_i}).
\end{equation}

where ${A_i}$ is the adjacency matrix of the data in the $i$-th client. ${h_i}$ is the output embedding of the $i$-th client trained in the local GNNs ${f_\theta }(A,X)$. Then, the server $S$ returns the global embeddings after propagating the embeddings of the clients by forward propagation. The server can use $h_{global}$ to train models for the main task, while the adversary can use them to infer private data.

Taking the node classification task as an example, the prediction of server $S$ can be expressed as:

\begin{equation}
	\widehat Y = softmax ({W_l} \cdot \rho (...\rho ({W_0} \cdot {h_{global}}))).
\end{equation}

where $\widehat Y$ is the set of node prediction labels, $W$ is the set of training weights, and $\rho $ is an aggregation function, usually ReLU.

The training loss of VGFL can be expressed as:

\begin{equation}
	{L_{tra}} =  - \sum\limits_{i = 1}^{{V_{train}}} {\sum\limits_{k = 1}^{|L|} {{Y_{i{\rm{k}}}}\ln ({{\widehat Y}_{ik}})} }.
\end{equation}

where $|L|$ is denoted as the number of categories of labels and $Y$ is the set of true labels of nodes.

\subsection{Threat Model}\label{3.3}

\begin{figure*}[!ht]
	\centering
	\includegraphics[width=1\textwidth]{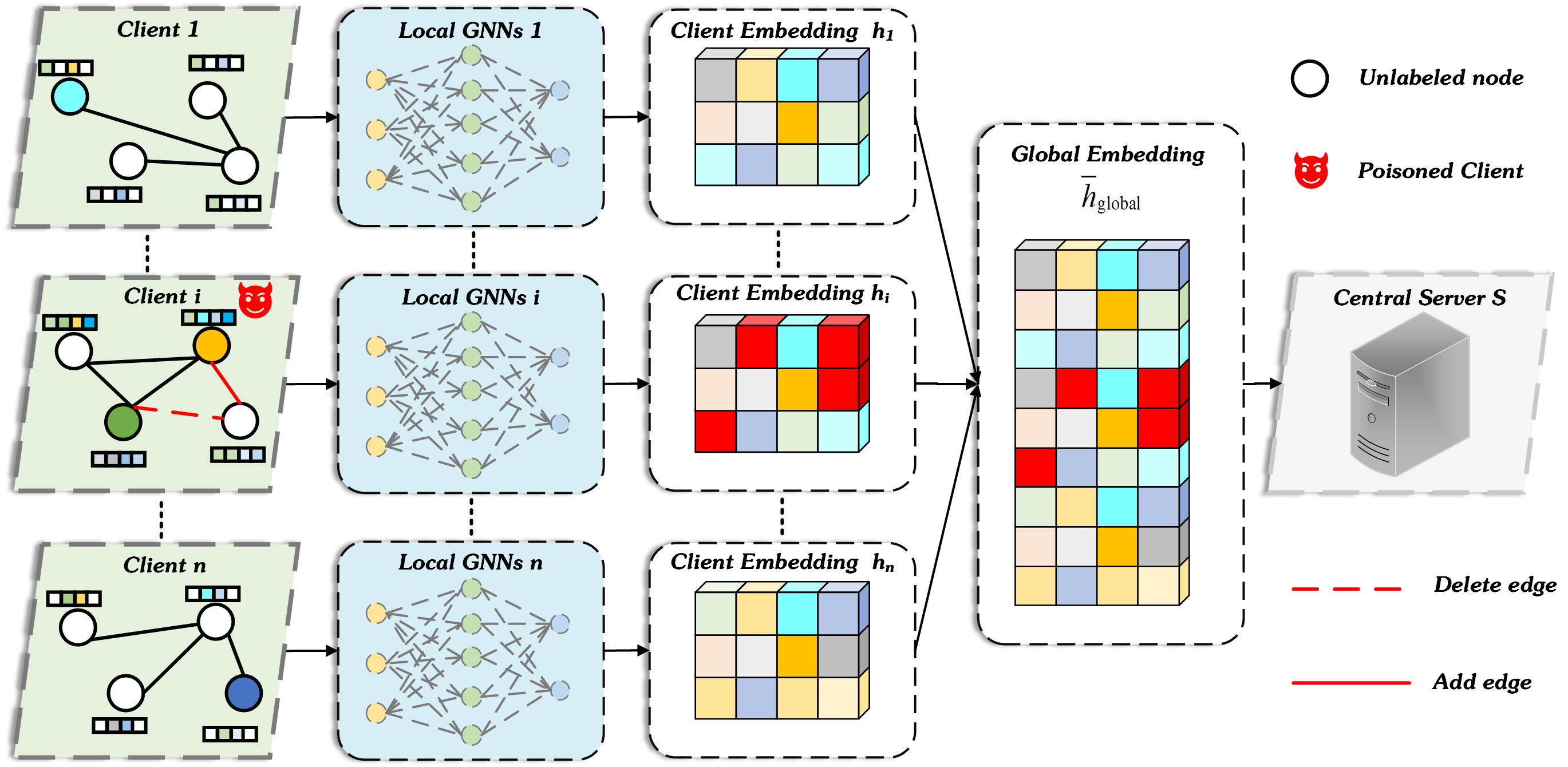}
	\caption{Diagram of the threat model.}\label{fig0}
\end{figure*}

We consider a common scenario where the central server model $S$ is trained collaboratively by multiple clients $C$ for downstream tasks, as shown in Fig. \ref{fig0}. The local GNNs are trained to get node embeddings, which are uploaded by the local client to the global server. 
In our model, the attacker does not know the structure of the vertical GFL, nor any parameter and gradient information of the server $S$. The attacker aims to generate malicious structural perturbations. When the clients are trained locally with poisoned data to generate poisoned embeddings, which are then uploaded into the central server $S$ to affect the final embeddings, which in turn leads to the output of incorrect results in the downstream tasks.

The aim of our proposed model is to accomplish structural attacks in the clients before Vertical GFL training. Then the clients are trained in local GNNs to generate embeddings with malicious information to be uploaded into the central server to maximize the training loss. It can be described by a mathematical formula as:

\begin{equation}
\begin{aligned}
& \operatorname{Max} \sum_{i \in V_{\text {test }}} L_{\text {tra }}\left(\hat{Y}_i, Y_i\right), \\
& \text { s.t. } \hat{Y}=\operatorname{soft} \max \left(W_l \cdot \rho\left(\ldots \rho\left(W_0 \cdot \bar{h}_{\text {global }}\right)\right)\right),\\ &\bar{h}_{\text {global }} \leftarrow \operatorname{CONCAT}\left(h_1, \ldots, \bar{h}_i, \ldots, h_K\right), \overline{h}_i=f_\theta^i\left(A_i, X_i\right) .
\end{aligned}
\end{equation}

where  ${\overline h _{global}}$ and ${\overline h _i}$ denote the global embedding and client embedding of poisoning, respectively.

\section{VGFL-SA Model}\label{S4}

\begin{figure*}[!h]
	\centering
	\includegraphics[width=1\textwidth]{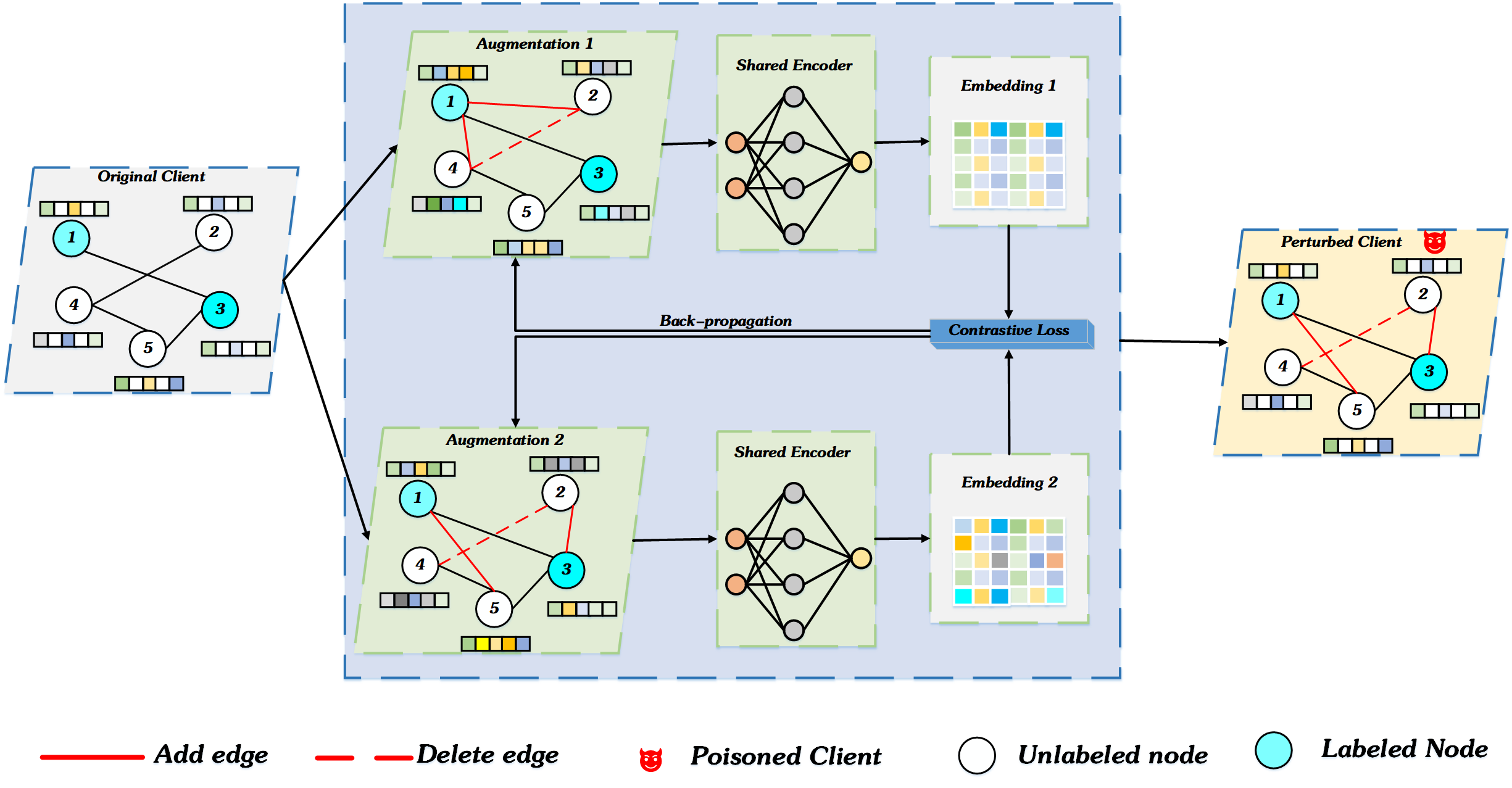}
	\caption{The general framework of VGFL-SA. Where, white nodes denote unlabeled nodes and blue nodes denote labeled nodes. VGFL-SA expansion to get two augmented views, using a shared encoder to get the embedding of the two augmented views. The gradient information of the neighbor matrix is obtained by comparative loss backpropagation, and the link with the largest absolute value of the gradient is flipped to obtain the perturbation graph of the poisoned client.}\label{fig2}
\end{figure*}

We consider a common case of VGFL where there are many clients working together to complete the training process for a global server.  Since only the node ID are shared among clients, for the graph structure is not shared. In other words,  the structure is different, which creates favorable conditions for the attack. In VGFL, some nodes in the local client are not labeled.  In the real world, using semi-supervised learning to implement the attack in the local client is difficult because the labels are hard to obtain.
We propose a structure attack against VGFL, referred to as VGFL-SA, which is an unsupervised attack model.
VGFL-SA aims to generate a perturbed graph by modifying the graph structure at a local client, then local GNNs are trained to get poisoned node embeddings, which are uploaded by the local client to the global server. The global server relies on embeddings uploaded by multiple clients, so embeddings containing poisoning information can lead to model performance in downstream tasks.
The overall framework of VGFL-SA is shown in Fig. \ref{fig2}.
First, the poisoned client is selected, two comparative views (Augmentation 1 and Augmentation 2) are obtained using data augmentation. The embedding of each  augmentation view is obtained using the shared graph encoder, and the difference between them is obtained through the comparative function, while the gradient of the neighbor matrix of the poisoned client is obtained by backpropagation. Finally, VGFL-SA flip the edge with the largest gradient to get the poisoned graph.

\subsection{Graph Augmentation}\label{4.1}

VGFL-SA uses graph contrastive learning to accomplish the unsupervised task, graph augmentation is essential for creating graph views, graph augmentation preserves and expands the topological and semantic information of the graph. VGFL-SA considers edge augmentation and node feature augmentation when augmenting the view, the specific augmentation methods are as follows.

\textbf{Edge Augmentation}. Consider an augmentation consisting of edge removal and edge addition. Specifically, given a client information ${G_i} = ({A_i},{X_i})$, where ${A_i}$ and ${X_i}$ denote the adjacency matrix and the feature matrix of the $i$-th client graph, respectively. Edge rewriting augments the structure by masking without considering node features.

\begin{equation}
	\begin{aligned}
		& A_i^j=A_i \cdot(1-I)+\left(1-A_i\right) \cdot I, \\
		& \text { s.t. } i \in[1, n], j \in[1,2], I_{x y} \sim\left\{\begin{array}{l}
			\text { Bernoulli }\left(p_{x j}^{-}\right), \text {if } A_{i, x y}=1 . \\
			\text { Bernoulli }\left(p_{x j}^{+}\right), \text {if } A_{i, x y}=0 .
		\end{array}\right.
	\end{aligned}\label{eq8}
\end{equation}

where $i$ denotes the number of clients and $j$ denotes the number of comparative views. $I$ is the indication matrix for rewiring, and ${A_{i,xy}}$ is denoted as the linking of the node $x$  and the node  $y$ in the $i$-th client.
$p_{xj}^ - $ and $p_{xj}^ + $ denote the probability of deleting and adding links to node $i$ and node $j$, respectively.
The indication matrix $I$ has only adding and deleting edges taking values, so we use the Bernoulli function to generate $I$, where to add the linking edges between node $i$ and node $j$, then ${I_{xy}} = 1$, otherwise ${I_{xy}} = 0$. 
Thus, ${A_i} \cdot (1 - I)$ denotes the delete edge operation and $(1 - {A_i}) \cdot I$ denotes the add edge operation in Eq. \ref{eq8}.
Note that in the indication matrix $I$, when there is a link between node $x$ and node $y$, there is a deletion probability $p_{xj}^ - $. Conversely, when there is no link between node $x$ and node $y$ , there is a deletion probability $p_{xj}^ + $.

The results of reference \cite{WOS:000749552300022} indicate that the larger the node degree value, the more information it contains, and that rewiring at such nodes tends to be more efficient. Inspired by this, the probability $p$ of a connected edge of Eq. \ref{eq8} is related to the node degree.
Suppose that in the $i$-th client, node $x$ and node $y$ have an edge $e$ whose edge importance can be expressed as:

\begin{equation}
	{S_{i,e}} = \log ({d_{i,x}} + {d_{i,y}}).
\end{equation}

where ${d_{i,x}}$ and ${d_{i,y}}$ are the node degrees of node $x$ and node $y$, respectively. Assuming that the maximum connected edge probability is ${s_{i,\max }}$ and the average connected edge probability is ${s_{i,avg}}$ in the $i$-th client, the deletion edge probability can be expressed as:
\begin{equation}
	p_e^ -  = \min (\frac{{{s_{i,\max }} - {s_e}}}{{{s_{i,\max }} - {s_{i,avg}}}},1).
\end{equation}

Similarly, the additive edge probability can be obtained as:

\begin{equation}\label{eq11}
	p_e^{+}=\min \left(\frac{s_e-s_{i, \min }}{s_{i, a v g}-s_{i, \min }}, 1\right).
\end{equation} 

where ${s_{i,\min }}$ is the minimum connected edge probability in the  $i$-th  client.

\textbf{Node Feature Augmentation}. In VGFL, different client nodes have only partial features and do not share all the features. When the number of clients is very large, the features owned by the nodes may be more sparse. Therefore, arbitrary deletion of features can mislead GNNs to understand the semantic information of nodes. Furthermore, assuming that certain node features are exclusive to a specific client, random augmentation of the node may lead to the generation of features that should not be present, thereby altering the semantic information of the node and preventing the optimal perturbation from being generated.
VGFL-SA considers a way of augmenting the node without rewriting its features, i.e., Feature Shuffling. 

The main idea of feature shuffling is to mix and shuffle the set of node features, which does not change the number of features and the distribution of the feature space.

Given a client graph, feature shuffling operates as follows:

\begin{equation} \label{eq12}
	X_i^j = {X_i}[idx,:], \quad {\rm{ }}s.t.{\rm{ }}  \quad i \in [1,n], {\rm{ }}j \in [1,2],idx = random(|V|).
\end{equation}

where $X_i^j$ is denoted as the $j$-th view of the $i$-th client and $Random (|V|)$ returns the node id number.

\subsection{Gradient Attack}\label{4.2}
Our goal is to perform graph contrastive learning using augmented views, using the differences between the augmented views as supervised signals, and then obtain the gradient of the adjacency matrix by contrastive learning. The performance of embedding is reduced by modifying the client links using gradient modification rules.

The above process can be described as:

\begin{equation}\label{eq13}
	\begin{aligned}
		& \max L\left(f_{\theta^*}\left(\mathrm{~A}_i^1, \mathrm{X}_i^1\right), f_{\theta^*}\left(\mathrm{~A}_i^2, \mathrm{X}_i^2\right)\right), \\
		& \text { s.t. } \theta^*=\arg \min {L}\left(f_\theta\left(\mathrm{A}_i^1, \mathrm{X}_i^1\right), f_\theta\left(\mathrm{A}_i^2, \mathrm{X}_i^2\right)\right),\\
		& \left\|\mathrm{A}-\mathrm{A}^{\prime}\right\| \leq \Delta, i \in[1, 2] .
	\end{aligned}
\end{equation}

where $f $ denotes a shared encoder, $\theta $ is a learnable parameter, and $i$ is the number of augmented views, set to 2. To ensure invisible rows for the attack, we set the attack budget $\Delta$.

Backpropagation of the Eq. \ref{eq13} to  obtain the gradient of the adjacency matrix. In GNNs, the value of the gradient reflects the sensitivity of the model to each node or edge in the input graph. Specifically, edges with larger absolute values of the gradient usually have a greater impact on the model's output. 
Deleting these edges effectively interferes with the model's learning process, leading to more significant performance degradation.

The gradient modification rule can be defined as:

\begin{equation}\label{eq14}
	\left\{\begin{array}{l}
		
		\text { Add } \mathrm{e}_{i, x y} \text {, s.t., } \nabla_{i, x y}>0, \mathrm{~A}_{i, x y}=0,\left|\nabla_{i, x y }\right|=\max \left|\nabla_i\right|, i \in[1, n] . \\
		\text { Delete } \mathrm{e}_{i, x y}, \text { s.t., } \nabla_{i, x y}<0, \mathrm{~A}_{i, x y}=1,\left|\nabla_{i, x y}\right|=\min \left|\nabla_{i,}\right|, i \in[1, n] .
	\end{array}\right.
\end{equation}

where ${\nabla _{i,xy}}$ denotes the gradient value of the adjacency matrix of node $x$ and node $y$ in the $i$-th client.

The core idea of Eq. \ref{eq14} is to add and delete edges based on the gradient, adding edges with the largest positive gradient and deleting edges with the smallest negative gradient. Deleting significant edges (i.e. edges with large gradient values) this can lead to maximizing the loss function of the model and can directly affect the propagation of information in the model, causing significant changes in the output.

In graph contrastive learning, GCN is often used in differentiable encoders, where the gradient of the augmented graph adjacency matrix can be computed by the following equation:

\begin{equation}\label{eq15}
	\left\{\begin{aligned}
		\nabla_i^1 & =\frac{\partial L}{\partial A_i^1}=\frac{\partial L}{\partial f\left(A_i^1, X_i^1\right)} \cdot \frac{\partial f\left(A_i^1, X_i^1\right)}{\partial L} , \\
		\nabla_i^2 & =\frac{\partial L}{\partial A_i^2}=\frac{\partial L}{\partial f\left(A_i^2, X_i^2\right)} \cdot \frac{\partial f\left(A_i^2, X_i^2\right)}{\partial L} ,
	\end{aligned} \text { s.t. } i \in[1,2].\right.
\end{equation}

Usually, the comparative learning encounters the problem of information bias, which leads to less than optimal performance. 
This is due to the fact that most of the works used randomness in performing the graph augmentation process, and the generated augmented graph has uncontrollable nature.
For example, some useless edges are added or removed randomly, which introduces noise and disturbs the true relationship between nodes. With this kind of augmentation, the model may incorrectly assume that some otherwise unrelated nodes are similar, which leads to inaccurate learned representations and reduces the generalization ability of the model.

Fortunately, VGFL-SA realizes that improper augmentation operations (e.g., excessive loss of information, introduction of extraneous noise) can lead to excessive structural differences between the augmented graph and the original graph, which means that the edges with the largest gradients in both views may not be the truly important edges in the original graph. VGFL-SA performs graph augmentation with respect to the graphs that are using the graph's topological properties (node degrees) to add or remove links, which makes the deviation of the two augmented graphs greatly reduced. However, VGFL-SA uses a randomized feature shuffling method for node feature generalization.
Although the final goal of VGFL-SA is to modify the links, we also need to consider the impact of feature bias on the results.

Therefore, we used a simple method to mitigate the bias introduced by random augmentation as follows:

\begin{equation}
	{\nabla _i} = \nabla _{_i}^1 + \nabla _{_i}^2,  \text { s.t. } i \in[1,2].
\end{equation} \label{eq16}

This method causes the larger (smaller)  gradients to be larger (smaller) in the augmented matrix. Finally, we use the gradient modification rule (Eq. \ref{eq14}) to select the edge with the largest absolute gradient and the correct gradient direction to flip. VGFL-SA selects only one edge in each round of the attack, so this method filters out many edges with large deviations and helps the attack find the best perturbation.

\subsection{Algorithm and Time Complexity}\label{4.3}
The pseudo-code of VGFL-SA is given in Algorithm \ref{alg1}.

\begin{algorithm}[ht!]  %ht!参数是调整算法在文章中的位置
	\renewcommand{\algorithmicrequire}{\textbf{Input:}}
	\renewcommand{\algorithmicensure}{\textbf{Output:}}
	\caption{VGFL-SA}  
	\label{alg1}
	\begin{algorithmic}[1] %每行显示行号
		\Require Graph dataset $G = (V,E,X)$, number of clients $K$, number of poisoned clients $ K'$, budget $ \Delta $, number of iterations $ T $
		\Require Perturbation dataset  $G' = (V,E^{\prime},X)$,  
		
		\State \textbf{Initialization:} Divide the client dataset $G_{i} = (V,E_{i},X_{i})$, pre-training GNNs model ${f_{{\Theta ^*}}}(A,X)$, select poisoned client set  $G' = (V,E',X)$
		
		\While { $||{A^{',t}} - A|| \le \Delta $}
		
		\For { $i=0 \le T $}
		\State Edge augmentation on poisoned clients through Eq. \ref{eq8} - Eq. \ref{eq11}
		\State Node feature augmentation on poisoned clients through Eq. \ref{eq12}
		\State Generate two comparison views $(\mathrm{~A}^1, \mathrm{X}^1),\left(\mathrm{~A}^2, \mathrm{X}^2\right)$
		\State Calculate the comparison loss $\nabla^1 $ and $\nabla^2 $ through Eq. \ref{eq13},  Eq. \ref{eq15} 
		\State To minimize the effects of bias, integrating gradient ${\nabla}$ through Eq. \ref{eq16}
		\EndFor
		
		\State Flip an edge with the largest absolute gradient to get the poisoned adjacency matrix $A^{',t}$ at time $t$ through Eq. \ref{eq14}
		
		\EndWhile

	\end{algorithmic}
\end{algorithm}

\textbf{Complexity Analysis.} 
VGFL-SA completes the attack before VGFL training and only needs to compute the client gradient information, so the time complexity is low. Specifically, VGFL-SA first needs to perform the augmentation and generalization operation to get the contrast views, and the time complexity can be denoted as $\mathcal{O}( T(n + d))$, $T$ is the number of iteration. Then, the contrast views shared encoder is trained and then the adjacency matrix gradient is obtained by backpropagation, and the time complexity can be denoted as ${\mathcal{O}(Tnd)}$.
When the graph is large, the time for shared encoder training is much larger than the time needed for view broadening, so the time complexity for VGFL-SA to have a single poisoned client is ${\mathcal{O}(Tnd)}$.
When the number of poisoned clients is $K'$, the time complexity of VGFL-SA is ${\mathcal{O}((T+K')nd)}$.

\section{Experiments}\label{S5}
\subsection{Datasets}\label{5.1}
For our experiments, we chose three commonly used benchmark citation datasets (Cora, Citeseer, PubMed). The overall information is listed in Table \ref{tab1}, where nodes represent papers and relationships between papers are described using edges. Specifically, the Cora dataset, created by Cornell University researchers, contains 2708 machine learning papers from the literature database, with each paper represented by a bag-of-words model of 1433 dimensions divided into seven domains, i.e., there are a total of 7 labels \cite{cora}. The Citeseer dataset, created by Cornell University researchers from the Citeseer Digital Library randomly selected papers, containing 3327 scientific papers, each represented by a bag-of-words model with 3703 dimensions, with 6 types of label \cite{cora}.  The PubMed dataset, created by the National Institutes of Health, contains 19717 biomedical papers, with 500 dimensions per node, with 3 labels \cite{DBLP:journals/aim/SenNBGGE08}.

\begin{table}[h]
	\begin{center}
		\caption{ Three commonly used datasets which are often used to evaluate the performance of graph attacks. Each graph is undirected and has no isolated nodes.}\label{tab1}%
		\begin{tabular}{c|cccc}
			\toprule
			Datasets & $\#$ Nodes & $\#$ Edges & $\#$ Features &  $\#$ Labels \\
			\midrule
			Cora     & 2708    & 5429    & 1433       & 7       \\
			Citeseer & 3327    & 4732   &  2879       & 6       \\
			PubMed & 19717    & 44325    & 500       & 3       \\
			\midrule
		\end{tabular}
	\end{center}
\end{table}

\subsection{Baselines}\label{5.2}
There are fewer current unsupervised model-based attacks, and we also selected some representative semi-supervised attack models. Specifically, unsupervised attacks are Random, UNEAttack \cite{UNEAttack}. Semi-supervised attacks are DICE \cite{DICE}, PGD \cite{PDG}, MinMax \cite{MinMax}, Metattack \cite{Metattack}, and Graph-Fraudster \cite{GraphFraudster}. Semi-supervised attacks use node labels as key information, unsupervised attacks do not use node labels. Therefore, semi-supervised attack models tend to perform better than unsupervised attack models. Fortunately, experimental results demonstrate that under some metrics, VGFL-SA outperforms some unsupervised attacks.

\textbf{Unsupervised Attacks:}

\textbf{Random}:  Random attack does not rely on a deep analysis of the graph, nor does it focus on the weights of the edges. It simply perturbs the edges by randomly modifying.

\textbf{UNEAttack} \cite{UNEAttack}: UNEAttack is the first unsupervised attack against graph adversarial attacks, which uses the theory of feature optima to disrupt node embeddings to degrade the performance of GNNs.

\textbf{Semi-supervised Attacks:}

\textbf{PGD} \cite{PDG}: PGD proposes a projected gradient descent based attack against a predefined model of GNNs (poisoning attack) using node labeling information.

\textbf{MinMax} \cite{MinMax}: MinMax uses node labeling information to propose a gradient-based attack against retrainable GNNs models (evasion attacks).

\textbf{Metattack} \cite{Metattack}: Metattack uses meta-learning algorithms for hyper-parameter optimization, with the core idea of optimizing input graph data as hyper-parameters for poisoning attacks in a black-box setting. 

\textbf{DICE} \cite{DICE}: DICE solves the network centrality metrics through heuristics and modifies the graph structure to accomplish the attack.

\textbf{Graph-Fraudster} \cite{GraphFraudster}: Graph-Fraudster is the first attack proposed for graph federated learning, which uses the gradient information of sum nodes to produce adversarial perturbations by adding noise to the global node embedding. Graph-Fraudster is an algorithm based on a targeting attack, i.e., it focuses on the classification of certain nodes. In this paper, we extend it to non-targeted attacks, trained with the perturbation graph generated by the attack, focusing on the classification situation of global nodes.

\subsection{Local GNNs}\label{5.3}

We use three commonly used GNNs to evaluate the effectiveness and transferability of VGFL-SA as detailed below:

\textbf{Graph Neural Network (GCN)} \cite{GCN}:  GCN is a deep learning model for graph-structured data. GCN achieves the aggregation and updating of node features by performing convolutional operations on node features using the adjacency matrix of the graph. 

\textbf{Graph Attention Neural Network (GAT)} \cite{GAT}: GAT introduces a self-attention mechanism that enables different weights to be dynamically assigned to neighboring nodes during feature aggregation. 

\textbf{Robust GCN} \cite{Robust}: Robust GCN aims to improve the robustness of graph neural networks to noise and perturbations. Robust GCN enhances the model's resistance to input perturbations by introducing regularization techniques and robustness mechanisms.

\subsection{Parameter Settings}\label{5.4}

In the local GNNs, the number of layers are set to 2 and the hidden layer output is 32 dimensions. In GAT, the number of attention heads is set to 2. The GNNs activation function is ReLU. Vertical GFL is optimized using Adam with a learning rate of 0.001. During training, we randomly divide the nodes into 10$\%$/10$\%$/80$\%$ of the train/test/validation sets. We run each experiment 10 times and report the average. For client data division, we use 5 clients, where the data ratio of clients 1 is 0.2. The number of poisoned clients is 2 by default. In graph contrastive learning, the number of contrastive views is set to 2, and 2 layers of GCN are used as encoders. To ensure the stealthiness of the attack, we set the attack budget to $\Delta  = \alpha m$, where $m$ is the number of edges and $\alpha $ is the budget factor, which is set to 0.1 by default.  

Our experimental environment consists of Xeon(R) Gold 6130 (CPU), Tesla V100 32 GiB (GPU), and 25 GiB memory.

\subsection{Assessment of Metrics}\label{5.5}
In order to evaluate the performance of VGFL-SA in detail, we use Accuracy, Precision, Recall, F1-Score, MAE and Log Loss metrics in our experiments. The details are as follows:

\begin{equation}
{\rm{Accuracy = }}\frac{{TP + TN}}{{TP + FN + FP + TN}}.
\end{equation}

\begin{equation}
{\rm{Precision = }}\frac{{TP}}{{TP + FP}}.
\end{equation}

\begin{equation}
{\rm{Recall = }}\frac{{TP}}{{TP + FN}}.
\end{equation}

\begin{equation}
{\rm{F1 = 2}} \times \frac{{{\rm{Precision}} \times {\rm{Recall}}}}{{{\rm{Precision + }}{\mathop{\rm Re}\nolimits} {\rm{cal}}l}}.
\end{equation}

\begin{equation}
\mathrm{MAE}=\frac{1}{n} \sum_{i=1}^n\left|\hat{y}_i-y_i\right|.
\end{equation}

\begin{equation}
Log{\rm{ }}Loss =  - \frac{1}{n}\sum\limits_{i = 1}^N {\sum\limits_{j = 1}^{|L|} {{y_{ij}}} \log {{\widehat y}_{ij}}}.
\end{equation}

Where, TP is the accurate optimistic predictions, FN is the false negative predictions, FP is the false positive predictions, and TN is the true negative predictions \cite{WOS:001294329300001}. $\hat{y}_i$ is the predicted value of node i and $y_i$ is the true value.
$n$ represents the number of samples, ${|L|}$ is the number of labels, the actual label of sample $i$ is (taking the value 0 or 1, indicating whether sample $i$ belongs to the $j$-th label), and the probability predicted by the model is ${\widehat y_{ij}}$ ( $0 \leq {\widehat y_{ij}} \leq 1$ indicating the probability that sample $i$ belongs to the $j$-th label).

\subsection{VGFL-SA Attack Performance in Accuracy}\label{5.4.1}

\begin{table*}[!ht]
	\caption{Accuracy ($\%$) results for different attacks in Vertical GFL (5 clients) where attacks are categorized into semi-supervised learning and unsupervised learning. Lower accuracy indicates better performance. The best results in both semi-supervised and unsupervised attacks are in bold. }\label{tab2}%
	\begin{center}
		\resizebox{\linewidth}{!}{
			\begin{tabular}{c|c|c|ccccc|ccc}
				\toprule
				&     & & \multicolumn{5}{c|}{Semi-supervised Attacks} & \multicolumn{3}{c}{Unsupervised Attacks}  \\
				\midrule
				GNNs & Datasets  & Clean &Metattack &PGD & MinMax  & DICE & Graph-Fraudster&Random &UNEAttack&VGFL-SA\\
				\midrule
				\multirow{3}*{GCN} &Cora &66.1&58.1&59.2&58.7&58.3&\textbf{57.8}&64.4&61.6&\textbf{59.9}\\
				&Citeseer & 62.7&\textbf{53.3}&	54.2&	54.8&	53.8	&53.7&	61.0&	58.0	&\textbf{56.6}\\
				&PubMed&67.2&\textbf{60.8}&	62.5&	61.7&	61.9&	61.5&	66.2&	64.1&	\textbf{61.8}\\
				\midrule
				\multirow{3}*{GAT} &Cora &66.8&	58.2&	59.6&	58.7&	58.9&\textbf{58.0}&	66.6&	61.5&\textbf{60.1}\\
				&Citeseer & 63.1& 54.6	&55.1&	55.8&	54.1	&\textbf{53.3}&	61.2&	58.3&	\textbf{56.4}\\
				&PubMed&66.8	&\textbf{60.2}&	61.2&	62.5&	61.4	&61.0&	65.1&	63.4&	\textbf{61.1}\\
				\midrule
				\multirow{3}*{Robust GCN} &Cora &65.4&	57.6&	58.4&	58.6&	57.8&	\textbf{57.3}&	63.4&	61.5&	\textbf{58.9}\\
				&Citeseer & 62.8&52.6&	54.4&	53.5&	53.4&\textbf{52.0}&	61.1&	58.1&\textbf{56.7}\\
				&PubMed&67.7&61.5& 62.1&	62.1&	62.5	&\textbf{61.3}&	65.2	&63.4&\textbf{62.0}\\
				\midrule
		\end{tabular}}
	\end{center}
\end{table*}

In this section, we examine the node classification results for several types of attacks on three common datasets, the results as shown in Table \ref{tab2}.

All the attacks shown are able to harm the performance of Vertical GFL as shown in Table \ref{tab2}, i.e., the attacks are able to cause a decrease in the accuracy of node classification. Specifically, the attack performance of semi-supervised learning tends to be better than the performance of unsupervised attacks. For example, using GCN as the victim model, the semi-supervised learning-based Metattack achieves the best F1-Score of 59.4$\%$ in PubMed, and our proposed unsupervised learning-based VGFL-SA achieves a performance of 60.6$\%$. 
This result is within our expectation because semi-supervised models usually outperform unsupervised models in graph deep learning. Semi-supervised learning makes use of partially labeled data, and this labeling information can help the model better learn the structure and features of the data. Attacks based on semi-supervised models are good at capturing the relationships between nodes, and can more accurately identify the key perturbations affecting GNNs through the information of labeled nodes. In Vertical GFL, when many client nodes exist unlabeled, semi-supervised learning-based attacks cannot proceed successfully, and the attacker can only use unsupervised model-based attacks to explore the robustness.  In particular, in PubMed, VGFL-SA outperforms PGD, MinMax and DICE. Due to the fact that PubMed has the highest average degree. A large amount of structural information can be retained in the client, and the more useful information that VGFL-SA can get. The quality of the generated structural perturbations will be high, and thus the performance is excellent.

In unsupervised model-based attacks, our proposed VGFL-SA achieves optimal performance in various datasets. For example, in Citeseer, Random, UNEAttack and VGFL-SA Log Loss are 1.33, 1.98 and 2.21 when GNNs are modeled as GAT, respectively. Since Random modifies edges randomly and does not rely on any graph information, it is not easy to identify key edges in the graph and will not be used often in real attacks. UNEAttack uses DeepWalk to generate node embeddings and then obtains structural perturbations through singular value decomposition. Since DeepWalk can only use the node index information, it cannot use the feature information of the nodes, which means that the quality of the generated perturbations is poor. From the above results, it can be seen that in unsupervised learning based attacks, our proposed model can generate structures with stronger perturbations in the same budget.

\subsection{VGFL-SA Attack Performance in Other Metrics}

\begin{table*}[!ht]
	\caption{Results for several types of attacks among other metrics. For Precision  ($\%$), recall  ($\%$) and F1-Score ($\%$),  the lower the attacker score, the better the attack performance. For MAE and Log Loss, the higher the attacker score, the better the attack performance. In both attacks, the optimal results are bolded in bold. }\label{tab3}%
	\begin{center}
		\resizebox{\linewidth}{!}{
			\begin{tabular}{c|c|ccccc|ccccc|ccccc}
				\toprule
				&     &   \multicolumn{5}{c|}{GCN}&  \multicolumn{5}{c|}{GAT} &  \multicolumn{5}{c}{Robust GCN}\\
				\midrule
				&     & Precision&	Recall&	F1-Score&	MAE	& Log Loss&Precision&	Recall&	F1-Score&	MAE	& Log Loss  &Precision&	Recall&	F1-Score&	MAE	& Log Loss   \\
				\midrule
				\multirow{10}*{Cora}& Clean	&65.2&	65.5&	64.4&	0.10&	1.22&66.0&	65.6&	66.2&	0.11&	1.15&65.2&	66.5&	66.4&	0.11&	1.18\\
				&Metattack	&58.0&58.7&	57.6&{0.20}&{2.32}&{58.9}&	{58.4}&	{57.8}&	{0.21}&	{2.30}&\textbf{58.0}&	{58.1}&{	57.4}&{	0.18}&	{2.33}\\
				&	PGD&	59.4&	59.8&	58.0&	0.18&	2.29&59.1&	59.2&	58.7&	0.18&	2.30&59.2&	58.6&	59.9&	0.18&	2.20\\
				&MinMax&	58.6&	58.0&	57.8&	0.19&	2.32&59.0&	59.8&	59.2&	0.18&	2.32&59.5&	59.1&	58.9&	0.17&	2.19\\
				&	DICE&	58.1&	59.2&	58.5&	0.18&	2.19&59.4&	58.1&	58.6&	0.18&	2.18&58.5&	59.6&	58.8&	0.17&	2.14\\
				&	Graph-Fraudster&\textbf{57.8}&	\textbf{56.6}&	\textbf{56.7}&	\textbf{0.23}&	\textbf{2.40}&\textbf{56.5}&	\textbf{55.8}&	\textbf{53.9}&	\textbf{0.24}&	\textbf{2.38}&\textbf{56.3}&	\textbf{58.0}&	\textbf{57.4}&\textbf{0.24}&	\textbf{2.36}\\ \cline{2-17}
				
				&	Random&	64.2&	64.6&	64.5&	0.14&	1.45&64.4&	63.8&	63.9&	0.13&	1.42&64.4&	64.0&	63.9&	0.13&	1.40\\
				&	UNEAttack&	61.4&	62.5&	61.3&	0.15&	1.80&60.5&	61.0&	61.1&	0.15&	1.77&61.8&	62.0&	61.4&	0.15&	1.71\\
				&	VGFL-SA	&\textbf{59.7}	&\textbf{59.2}&\textbf{	58.1}&	\textbf{0.18}&	\textbf{2.31}&\textbf{59.5}&	\textbf{59.0}&\textbf{59.1}&	\textbf{0.17}&	\textbf{2.32}&\textbf{59.6}&	\textbf{60.5}&	\textbf{59.3}&	\textbf{0.17}&	\textbf{2.30}\\
				\midrule
				\multirow{10}*{Citeseer}& Clean	&62.4&	62.6&	63.1&	0.13&	1.17&63.2&	63.1&	63.0&	0.13&	1.15&63.0&	62.2&	62.5&	0.13&	1.13\\
				&Metattack	&\textbf{52.4}&	\textbf{53.6}&	\textbf{52.2}&	\textbf{0.22}&	\textbf{2.49}&\textbf{53.0}&	\textbf{53.4}&	\textbf{52.5}&	\textbf{0.21}&	\textbf{2.40}&\textbf{52.0}&	{53.1}&	52.7&	\textbf{0.21}&	\textbf{2.38}\\
				&	PGD&	54.1&	54.0&	54.4&	0.20&	2.25&54.3&	55.4&	54.2&	0.19&	2.21&54.8&	53.5&	53.4&	0.19&	2.18\\
				&MinMax&	54.5&	55.4&	54.1&	0.20&	2.27&55.6&	56.4&	55.1&	0.19&	2.19&54.2&	53.2&	53.4&	0.18&	2.16\\
				&	DICE&	53.4&	53.5&	53.4&	0.19&	2.18&54.5&	53.4&	53.7&	0.18&	2.20&53.4&	54.2&	53.0&	0.18&	2.15\\
				&	Graph-Fraudster&	54.1&	54.0&	54.4&	0.20&	2.38&54.4&	54.4&	53.2&	0.18&	2.37&52.2&	\textbf{53.0}&	\textbf{52.4}&	0.17&	2.36\\ \cline{2-17}
				
				&	Random&	61.5&	60.7&	61.6&	0.14&	1.33&61.3&	62.3&	61.5&	0.14&	1.33&61.2&	60.8&	60.9&	0.14&	1.30\\
				&	UNEAttack&	58.3&	58.2&	57.7&	0.17&	2.07&58.4&	57.2&	57.3&	0.17&	1.98&57.5&	57.2&	57.7&	0.16&	1.89\\
				&	VGFL-SA	&	\textbf{54.8}&	\textbf{55.7}&	\textbf{55.9}&	\textbf{0.19}&	\textbf{2.20}&\textbf{55.2}&	\textbf{56.7}&	\textbf{54.4}&	\textbf{0.19}&	\textbf{2.21}&\textbf{54.8}&	\textbf{55.2}&	\textbf{54.3}&	\textbf{0.18}&	\textbf{2.23}\\
				\midrule
				\multirow{10}*{PubMed}& Clean	&67.1&	67.7&	67.8&	0.15&	1.02&67.9&	67.5&	67.4&	0.15&	1.01&66.8&	67.3&	67.5&	0.15&	0.97\\
				&Metattack	&\textbf{60.6}&	\textbf{59.0}&	\textbf{59.4}&	\textbf{0.21}&	\textbf{3.35}&\textbf{59.5}&	\textbf{60.9}&	\textbf{60.6}&	\textbf{0.20}&	\textbf{3.30}&\textbf{60.9}&	\textbf{60.2}&	\textbf{59.2}&	\textbf{0.20}&	\textbf{3.28}\\
				&	PGD&	61.5&	61.4&	61.9&	0.20&	3.33&62.2&	61.4&	61.7&	0.19&	3.15&61.8&	62.3&	61.2&	0.19&	3.16\\
				&MinMax&	60.8&61.9&	61.7&	0.18&	3.18&61.4&	62.1&	61.8&	0.18&	3.18&62.6&	61.8&	61.4&	0.19&	3.14\\
				&	DICE&	61.5&	62.4&	61.0&	0.18&	3.25&61.8&	62.1&	61.5&	0.18&	3.14&62.4&	61.3&	61.6&	0.18&	3.16\\
				&	Graph-Fraudster&61.4&	61.6&	61.5&	0.19&	3.23&61.5&	62.0&	61.4&	0.18&	3.25&61.3&	62.3&	61.5&	0.18&	3.26\\ \cline{2-17}
				&	Random&65.3&	66.2&	65.0&	0.16&	1.65&66.3&	66.2&	65.8&	0.16&	1.62&66.4&	66.0&	65.8&	0.16&	1.60\\
				&	UNEAttack&62.4&	63.2&	62.7&	0.17&	2.81&63.2&	62.5&	63.6&	0.17&	2.72&62.8&	62.7&	62.4&	0.17&	2.70\\
				&	VGFL-SA	&\textbf{61.2}&	\textbf{62.5}&	\textbf{60.6}&	\textbf{0.20}&	\textbf{3.32}&\textbf{61.2}&	\textbf{61.5}&	\textbf{61.4}&	\textbf{0.19}&	\textbf{3.20}&\textbf{62.3}&	\textbf{61.4}&	\textbf{61.5}&	\textbf{0.19}&	\textbf{3.15}\\
				\midrule
		\end{tabular}}
	\end{center}
\end{table*}

To fully demonstrate the performance of VGFL-SA, we also conducted experiments in Precision, Recall, F1-Score, MAE and Log Loss metrics. In most cases, the results are the same as in the  Accuracy metric, i.e., Metattack performs best in semi-supervised learning-based attacks and VGFL-SA performs best in unsupervised learning-based attacks. However, in a few cases, our proposed model can outperform semi-supervised learning based attacks in some metrics. For example, in PubMed, when the victim model is GCN, the F1-Score of VGFL-SA is 59.6$\%$, and the Recall of PGD, MinMax, DICE, and Graph-Fraudster are 59.9$\%$, 60.7$\%$, 61.0$\%$, and 60.5$\%$, respectively. This indicates that VGFL-SA recognizes irrelevant noise in the graph and achieves the desired results without using node labeling generated perturbations.

In addition, we also verify the transferability of the attack. We transfer the victim model to GAT and RobustGCN and the results are shown in Table \ref{tab2}  and Table \ref{tab3}. Observations show that VGFL-SA achieves good performance with different GNNs. For example, the classification accuracy and Recall of VGFL-SA are $\{$ 56.7$\%$, 55.9$\%$$\}$ and $\{$ 55.4$\%$, 55.5$\%$$\}$ in GAT and RGCN when the dataset is Cora, respectively. In conclusion, our proposed model also has good transferability.

\subsection{Comparative Learning Components}\label{5.4.3}

\begin{table}[!ht]
	\caption{The effect of several comparative learning components on classification accuracy ($\%$), with the victim modeled as GCN, GAT and Robust GCN. VGFL-SA-V1 is a randomly augmented version of the VGFL-SA structure. VGFL-SA-V2 is a randomly augmented version of the VGFL-SA features, specifically, the features are randomly changed rather than transforming the location as in Feature Shuffling. VGFL-SA-V3 is a randomly augmented version of the VGFL-SA structure and features. In both attacks, the optimal results are bolded in bold. }\label{tab3.1}%
	\begin{center}
		\resizebox{\linewidth}{!}{
			\begin{tabular}{c|ccc|ccc|ccc}
				\toprule
				&\multicolumn{3}{c|}{GCN}&\multicolumn{3}{c|}{GAT}&\multicolumn{3}{c}{Robust GCN}\\
				&   Cora	& 	Citeseer	& 	PubMed   &   Cora	& 	Citeseer	& 	PubMed &   Cora	& 	Citeseer	& 	PubMed\\
				\midrule
				VGFL-SA-V1& 	60.8& 	57.3& 	62.6 & 60.9&57.3&61.6&59.8&57.7&62.5 \\
				VGFL-SA-V2& 	60.3& 	56.9& 	62.2 & 60.5&56.9&61.3&59.4&57.4&62.3  \\
				VGFL-SA-V3& 	61.5& 	58.2& 	63.5 & 61.6&57.9&62.3&60.7&58.5&63.6 \\
				VGFL-SA& 	\textbf{59.9}& 	\textbf{56.6}& 	\textbf{61.8}& \textbf{60.1}&\textbf{56.4}&\textbf{61.1}&\textbf{58.9}&\textbf{56.7}&\textbf{62.0} \\
				\midrule
		\end{tabular}}
	\end{center}
\end{table}

The experiments in this section investigate the effect of the comparative learning components on the performance of the attack, and the results are shown in Table  \ref{tab3.1}.  We find that the higher the degree of randomization, the worse the performance of the model in comparative learning. For example, VGFL-SA-V3 sets both structure and features to be randomly augmented, and it has the lowest classification accuracy in all datasets, with an average performance 2$\%$ lower than VGFL-SA. Moreover, structure randomization is not as effective as feature randomized augmentation. In other words,  the original graph structure information is a bit more important in the augmented graph. For example, in Cora, the classification performance of VGFL-SA-V1 with randomized augmentation of structure is 0.5 lower than that of VGFL-SA-V2 with randomized augmentation of features. This also verifies that attacking structure is more effective than attacking features under the same budget in graph adversarial learning. Finally, our proposed VGFL-SA augmentation method achieves the best performance among several classes of extended models, which indicates that our generated augmented graphs effectively retain the structural information and with the semantic information in the original graph. When generating perturbations, VGFL-SA identifies the most sensitive edges in the graph to the model predictions, enabling the perturbed graph to maximize the change in the model predictions.

\subsection{Attack Budget}\label{5.4.2}

\begin{figure}[hpt]
	\centering
	\includegraphics[width=1\textwidth]{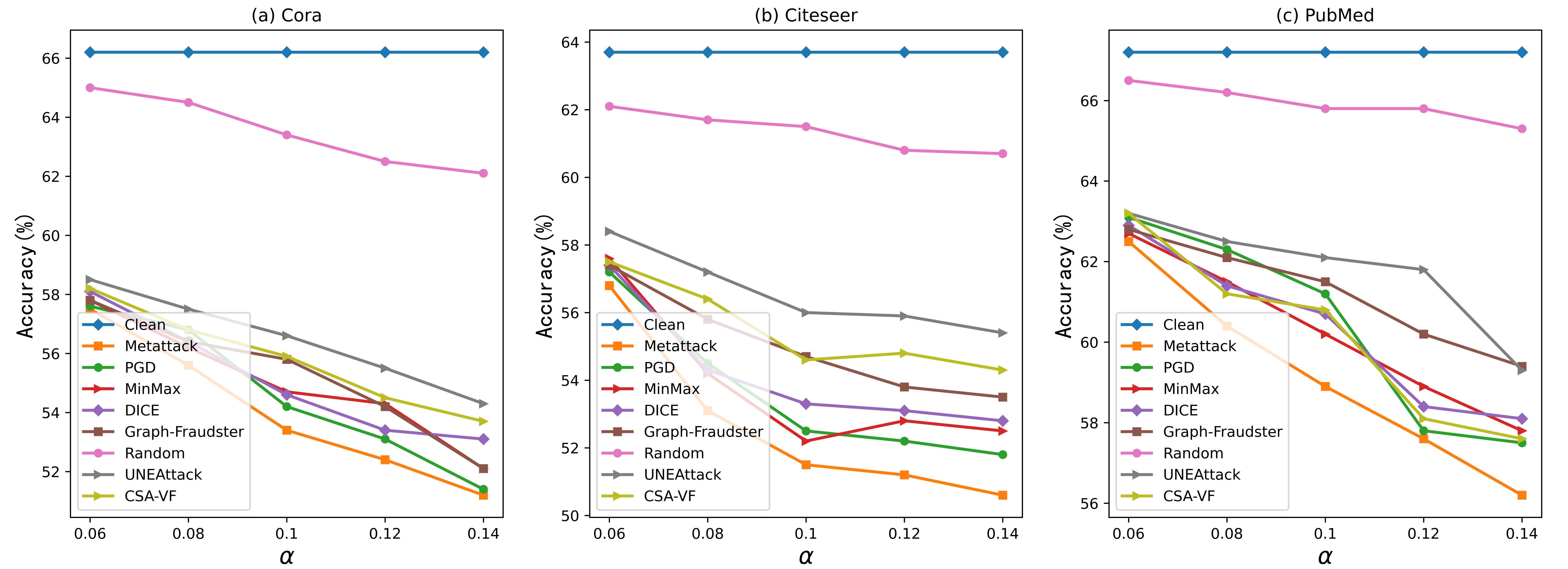}
	\caption{ Attack performance of several types of attacks under different attack budgets, $\alpha $ is the perturbation factor. The larger the $\alpha $, the larger the attack budget.}\label{fig3}
\end{figure}

The experiments in this section investigate the effect of attack performance on attack budget, and the results are shown in Fig. \ref{fig3}. We use the accuracy metric to measure the attack performance. For the validity of the experiment, the perturbation factors $\alpha $ are set to 0.06, 0.08, 0.1, 0.12 and 0.14.

From the results in Fig. \ref{fig3}, it can be seen that as the perturbation factor $\alpha $ increases, the node classification performance decreases, i.e., the attack performance is positively proportional to the attack budget. When the perturbation factor $\alpha $ increases, the higher the number of modified links, the more harmful information Vertical GFL learns, which eventually leads to incorrect decision making in downstream tasks. When the perturbation factor $\alpha $ is small, there is little difference in the performance of all attacks. As the perturbation factor   $\alpha $ increases, the difference in attack performance starts to increase. For example, in PubMed, when the perturbation factor $\alpha $ are 0.06 and 0.12, the performance of Metattack, Graph-Fraudster and VGFL-SA are $\{$62.5$\%$, 56.2$\%$$\}$, $\{$62.8$\%$, 56.4$\%$$\}$ and $\{$63.2$\%$, 57.6 $\%$$\}$, respectively. Comparing several types of attacks, it can be found that our proposed model achieves the best performance in the unsupervised learning model regardless of the perturbation rate. In particular, VGFL-SA (57.6$\%$) can outperform some semi-supervised models (MinMax: 57.8$\%$, DICE: 58.1$\%$) when the perturbation factor $\alpha $ is 1.2 in PubMed.

\subsection{Number of clients}\label{5.4.4}

\begin{figure}[!hpt]
	\centering
	\includegraphics[width=1\textwidth]{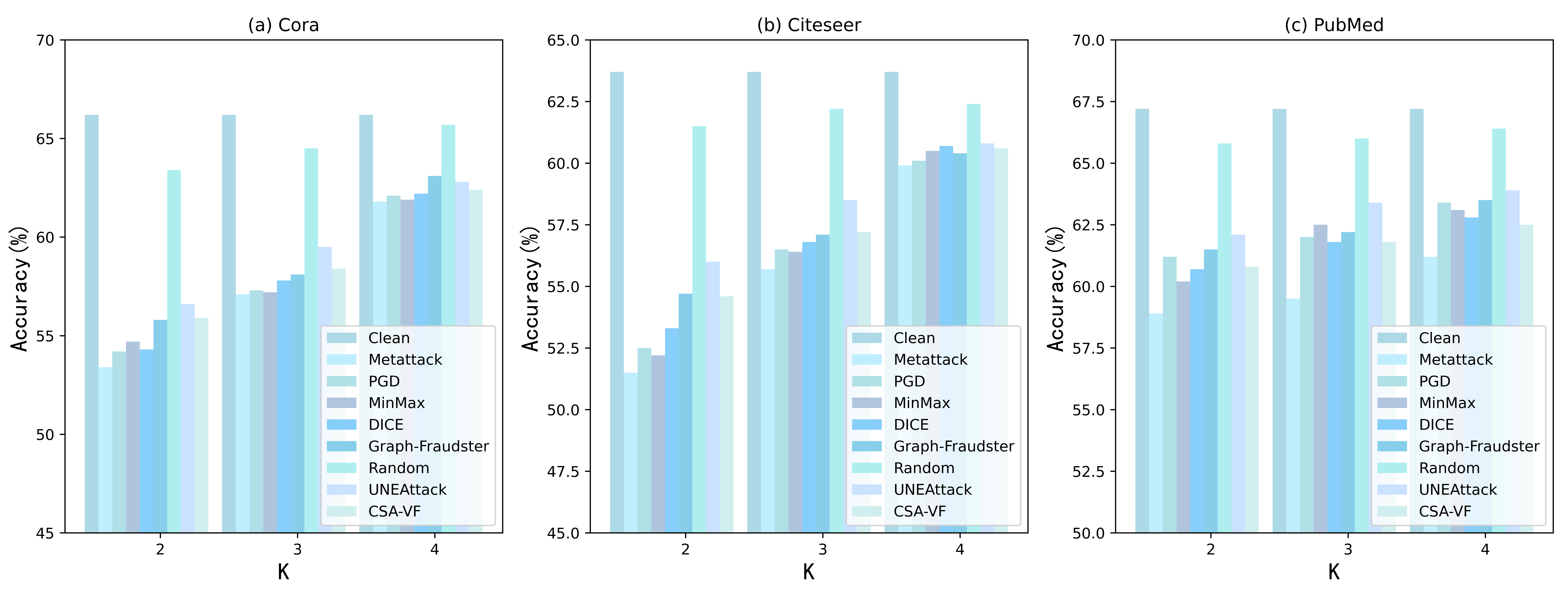}
	\caption{  Classification accuracy of VGFL-SA in multiple clients-one poisoned client, where $K$ is the number of clients and the number of poisoned clients is 1.}\label{fig4}
\end{figure}

\begin{figure}[!hpt]
	\centering
	\includegraphics[width=1\textwidth]{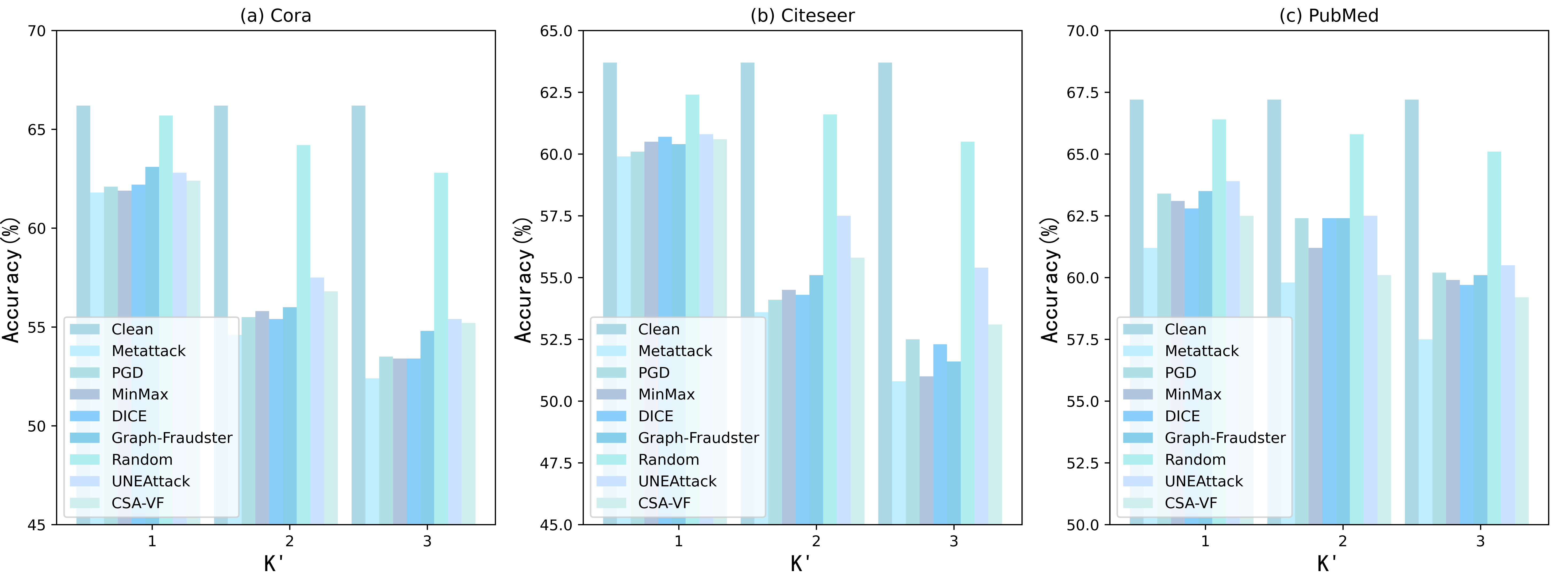}
	\caption{ Classification accuracy of VGFL-SA over multiple clients-multiple poisoned clients, where the number of all clients is 4 and $K'$ is the number of poisoned clients.}\label{fig5}
\end{figure}

In this section, the effect of the number of clients on the attack is investigated for the cases of multiple clients-one poisoned client and multiple clients-multiple poisoned clients, and the results are shown in Fig. \ref{fig4} and Fig. \ref{fig5}.

Fig. \ref{fig4} shows the classification performance of various attacks in the multiple client-one poisoned client case. In all datasets, as the number of multiple clients increases, the classification accuracy of the attacks also rises, i.e., the attack performance decreases.
For example, in PubMed, the classification accuracy of VGFL-SA is 60.8$\%$, 61.8$\%$ and 62.5$\%$ when the number of clients $K$ is 2, 3 and 4, respectively. This is because in Vertical GFL, clients usually do not share raw data among themselves, but share model information. As the number of clients $K$ increases, the data is dispersed more, and it is more difficult for the attacker to obtain enough information to construct an effective counter sample. In addition, when the number of clients is high, malicious information from poisoned clients may be canceled out, leading to more stable updates to the global model.

Fig. \ref{fig5} shows the performance of several types of attacks with different number of poisoned clients $K'$ among 4 clients. It can be observed that when the number of clients $K'$ is constant, the more poisoned clients cause higher damage to the Vertical GFL performance. For example, in Cora, when the number of poisoned clients  $K'$  is 1,2,3, the VGFL-SA classification accuracy is 62.4$\%$, 56.8$\%$, and 55.2$\%$, respectively. Under other attacks and datasets, the results are the same, i.e., as the number of poisoned clients increases, the performance of the attack improves. As the number of poisoned clients increases, the proportion of malicious messages in the central server increases and the results output is less accurate.

\subsection{Special case analysis-Proportion of poisoned clients' data}\label{5.4.5}

In this section, the impact of participant data percentage on VGFL-SA performance is investigated and the results are shown in Table \ref{tab61}.
In order to see a clear difference in the results, in this section we set the number of clients to 2, where the number of poisoned clients is 1 and the number of clean clients is 1.
Proportion of poisoned clients' data are set to 0.3, 0.4, 0.5, 0.6, 0.7, 0.8.

\begin{table}[!h]\label{tab61}
	\centering
	\resizebox{\linewidth}{!}{
		\begin{tabular}{cccccccc}
			\toprule
			& Proportion & Accuracy & Precision & Recall & F1-Score & MAE & Log Loss \\
			\midrule
			\multirow{6}{*}{Cora} & 0.3 & {58.5} & {58.4} & {56.1} & {56.5} & {0.16} & {1.85} \\
			& 0.4 & 57.1 & 57.2 & 54.0 & 54.8 & 0.16 & 1.96 \\
			& 0.5 & 55.9 & 0.54 & 0.53 & 0.53 & {0.17} & {2.08} \\
			& 0.6 & 55.5 & 56.1 & 53.6 & 53.4 & 0.19 & 2.15 \\
			& 0.7 & 54.7 & 55.2 & 53.1 & 53.2 & 0.20 & 2.54 \\
			& 0.8 & 53.5 & 54.2 & 52.8 & 52.8 & 0.21 & 2.63 \\
			\midrule
			\multirow{6}{*}{Citeseer} & 0.3 & 57.8 & 58.2 & 57.8 & 58.3 & 0.18 & 1.78 \\
			& 0.4 & 55.4 & 55.8 & 55.6 & 55.2 & 0.18 & 2.34 \\
			& 0.5 &{54.6} & {0.54} & {0.53} & {0.54} & {0.19} & {2.20} \\
			& 0.6 & 53.2 & 53.4 & 53.1 & 53.2 & 0.20 & 2.48 \\
			& 0.7 & 52.4 & 52.8 & 52.7 & 52.4 & 0.21 & 2.61 \\
			& 0.8 & 52.1 & 51.8 & 51.6 & 51.8 & 0.21 & 2.72 \\
			\midrule
			\multirow{6}{*}{PubMed} & 0.3 & 62.4 & 62.4 & 61.8 & 61.4 & 0.17 & 2.45 \\
			& 0.4 & 61.4 & 60.5 & 60.5 & 60.1 & 0.19 & 2.78 \\
			& 0.5 & {60.8} &{0.61} &{0.60} & {0.58} & {0.20} &{3.32} \\
			& 0.6 & 59.4 & 58.6 & 58.9 & 57.5 & 0.20 & 3.82 \\
			& 0.7 & 58.4 & 58.1 & 57.8 & 57.1 & 0.21 & 4.23 \\
			& 0.8 & 57.8 & 58.0 & 57.1 & 56.8 & 0.22 & 4.61 \\
			\bottomrule
	\end{tabular}}
	\caption{Performance metrics for different datasets and parameters.}
\end{table}

It is observed that as the percentage of poisoned client data increases, the performance of VGFL-SA improves under various metrics. For example, in Cora, when the proportion of poisoned client data is 0.3 and 0.8, the results of Accuracy, Precision, Recall and F1-Score metrics of VGFL-SA are $\{$58.5$\%$, 53.5$\%$$\}$, $\{$58.4$\%$,  54.2$\%$$\}$, $\{$56.1$\%$, 52.8$\%$$\}$ and $\{$56.5$\%$, 52.8$\%$$\}$, respectively. The smaller the Accuracy result, the better the attack performance. In MAE and Log Loss metrics results are $\{$0.16, 0.21$\}$, $\{$1.85, 2.63$\}$ respectively. The larger the MAE and Log Loss results, the better the attack performance. This is because in Vertical GFL, a client with a larger proportion of data has a greater impact on the central server. Therefore, when the poisoned client holds more data, the worse the quality of the generated node embedding and the worse the Vertical GFL performance.

\section{Conclusion}\label{Sec:conclusion}

In order to explore the robustness of VGFL, this paper proposes an unsupervised attack (VGFL-SA). VGFL-SA uses contrastive learning to complete the attack by modifying the client structure using only the graph structure and node features information. VGFL-SA solves the problem that the current attack methods can not be implemented in the unlabeled tagged environments. VGFL-SA firstly accesses the poisoned clients to augment the graph generating contrastive views. VGFL-SA uses node degree properties to generate edge augmentation, and use feature shuffling to generate feature augmentation. VGFL-SA then uses shared encoder backpropagation to obtain the gradient of the adjacency matrix and complete the attack by modifying the edges with the largest absolute value of the gradients. When the local clients complete the training to generate embeddings with poisoned information and then uploads them to the central server, the embeddings with poisoned information will lead to performance degradation of the central server in downstream tasks.
We have shown through extensive experiments that VGFL-SA can outperform various unsupervised comparison models under various evaluation metrics, and outperform semi-supervised comparison models in some cases.

We initially explores the robustness of graph federated learning in this work. In the future, we focus on achieving attack extensions through cross-domain attacks. For example, disruption of large-scale graph federation learning systems can be achieved by attacking data in one domain and affecting multiple domains related to it. In addition, we focus on distributed defense strategies suitable for graph federation learning. In particular, it remains a great challenge to effectively detect and defend against attacks from multiple participants without exposing the original data.

%Bibliography
\bibliographystyle{unsrt}  
\bibliography{references}

\end{document}